\pdfoutput=1
\documentclass[11pt]{article}

\usepackage{acl}
\usepackage{times}
\usepackage{latexsym}
\usepackage{graphicx}
\usepackage{subfigure}
\usepackage{enumitem}
\usepackage{adjustbox}
\usepackage[T1]{fontenc}
\usepackage[utf8]{inputenc}

\usepackage{microtype}

\newcommand{\OUT}[1]{}

\title{Reference Resolution and Context Change  in Multimodal Situated Dialogue for Exploring Data Visualizations}

\author{Abhinav Kumar \\
  University of Illinois Chicago \\
  \texttt{akumar34@uic.edu} \\\\
  \bf Jillian Aurisano \\
  University of Cincinnati \\
  \texttt{jillian.aurisano@uc.edu} \\\And
  Barbara Di Eugenio \\
  University of Illinois Chicago\\
  \texttt{bdieugen@uic.edu} \\\And
  Abari Bhattacharya \\
  University of Illinois Chicago \\
  \texttt{abhatt62@uic.edu}  \\\\
 \bf Andrew Johnson \\
  University of Illinois Chicago \\
  \texttt{ajohnson@uic.edu }
  }

\begin{document}
\maketitle
\begin{abstract}
Reference resolution, which aims to identify entities being referred to by a speaker, is more complex in real world settings:  new referents may be  created by processes the agents engage in and/or be salient only because they belong to the shared physical setting.  Our focus  is on resolving references to visualizations on a large screen display in multimodal dialogue; crucially, reference resolution is directly involved in the process of creating new visualizations.  We describe\OUT{our \OUT{low resource} multi-modal dialogue and} our annotations for \OUT{anaphoric} user references to visualizations appearing on a large screen via language and hand gesture and also new entity establishment, which results from executing the user request to create a new visualization. \OUT{to be constructed shortly after the user asks to construct them in conversation.} We also describe our reference resolution pipeline which relies on an information-state architecture to maintain dialogue context. \OUT{ on the current visualizations on the screen.}\OUT{and highlight the strengths and weaknesses}We report results on detecting and \OUT{anaphora} resolving references, effectiveness of contextual information on the model, and under-specified requests for creating visualizations. We also experiment with conventional CRF and \OUT{the state of the art} deep learning / transformer models (BiLSTM-CRF and BERT-CRF) for tagging \OUT{anaphoric} references in user utterance text. Our results show that transfer learning significantly boost performance of the deep learning methods, \OUT{compared to the baseline} although \OUT{conventional} CRF still out-performs them, suggesting that conventional methods may generalize better \OUT{for certain problems and domains} for low resource data.
\end{abstract}

\section{Introduction}
Conversation is understood in context.
\OUT{, in the simplest terms requiring interpretation and relating what the speaker is saying to \OUT{our own understanding and} knowledge about the world and specific domain.} \OUT{As such,} When  the world, whether real or simulated, can change because of the user's actions, new entities are created by the processes that change the world itself; in this context, reference resolution, which is tasked with linking what the user refers to with  objects in the world, is crucial  for  a dialogue system to \OUT{understand and} effectively respond to the user, including to create the new entities themselves. 

In this work, we discuss our approach to new entity establishment and reference resolution to deal with references to visualizations on a large screen display where  new visualizations are constantly being added and then  moved, opened/closed,  or even  removed. In our {\it CITY-CRIME-VIZ corpus},  (see Section~\ref{sec:corpus}),  the user may ask {\it $U_1$: can I see theft in the downtown area}, resulting in a first visualization Viz$_1$;  and then {\it $U_2$: can you show that graph by day of the week?}, which results in  a second visualization Viz$_2$. Viz$_2$ is  created by first resolving the referring expression {\it that graph} to Viz$_1$, and then generating the specifications for  Viz$_2$ by updating the specifications for  Viz$_1$ according to $U_2$. This is an example of {\it accommodating context change}, a notion first introduced by \cite{webber-baldwin1992} in their discussion of \OUT{introducing and referring to} new entities that are the results of physical processes as in cooking: {\it Mix the flour, butter and water. Knead {\underline the dough} until smooth and shiny.} In the  30 years  since \cite{webber-baldwin1992}, not much work has been done on how to accommodate the creation \OUT{(whether real or simulated)} of new entities (see \citep{wilson2016table}  for  documents  and \citep{li2016reference} for tutoring dialogues about programming), and none in the visualization domain; and  none that marries traditional information-state architecture with contemporary lexical representations and approaches as we do here. Additionally, our task is inherently multimodal, specifically as concerns deictic gestures. \OUT{\citep{navarretta2011anaphora,qu2008beyond,landragin2006visual,eisenstein2006gesture}.}

\OUT{In \cite{webber-baldwin1992}, the new entities that are introduced in the context are still described in text as results of processes. However, when the processes actually take place in a real or simulated world, then the user refers to objects not only through language, but also nonverbal cues such as gestures \citep{navarretta2011anaphora,qu2008beyond,landragin2006visual,eisenstein2006gesture}, eye gaze \citep{prasov2008s,iida2011multi,kim2017multi}, and haptic interfaces \citep{foster2008roles,chen2015roles}. The objects being referred to could be text entities introduced earlier in the discourse or  those in external environments, such as icons on a screen \citep{kehler2000cognitive}, ingredients for a recipe \citep{whitney2016interpreting}, and in the kitchen finding useful objects (e.g., \textit{"spoon"}) in specific locations (e.g., \textit{"drawer"}) \citep{chen2012co}. }

Our overall research goal is to build a conversational assistant to support  users  explore data visualization via multimodal interaction. We collected a  corpus of  interactions \textit{City-Crime-Vis} from  16 subjects tasked with forming effective police deployment strategies based on crime patterns discovered while exploring visualizations of our city's public data. The subjects  interacted with a Visualization Expert (VE) which they  knew to be a person  interacting remotely from a separate room. We have designed, developed and evaluated a first version of our assistant (References withheld); the reference resolution work we describe here is part of a second version of the assistant.

We highlight our contributions as follows. (1) We believe we are the first to code for visualization references in a large screen environment supporting multiple visualizations at a time. We tagged \OUT{in particular visualization} referring noun phrases (NPs), data attributes (slot fillers)  in these NPs, and   co-occurring pointing gestures 
\OUT{. We also tagged certain words in the construction request (when the visualization is introduced to the discourse) as slot fillers. These correspond to some subset of the data attributes displayed by a visualization (} Because the user rarely fully specifies the visualization to be created,  we can use the slot fillers as components of semantic structures that are used both to find the referent visualization, and to create the specification for the new one. 
\OUT{represent visualizations created on the fly via semantic structures  and form a similar structure for referring expressions for the purpose of selecting the target visualization the user is most likely referring to on the screen.} (2) 
\OUT{As far as we know,} Existing visualization software have applied rule-based grammars for detecting references in text, which are difficult to scale, including to the different forms these references can take (pronouns, deictics, full NPs).
\OUT{however it would be difficult to scale to all the possible varying scenarios that free-forming NL presents in our case. The user oftentimes would refer to anything in our environment not just visualizations, such as objects within visualization, visualization properties, and so on.  Also references in the utterances were phrased in a multitude of ways by the user, taking on different forms including 3rd person pronouns, demonstratives, and deictics.} Alternatively, we 
\OUT{propose a machine learning (ML) based approach} use conditional random field (CRF) to detect referential expressions, \OUT{tag the best sequence of referring expression (RE) labels to the utterance text,} achieving an F1 score of 61.6\% on our corpus data. (3) Furthermore \OUT{given deep learning models have achieved state of the art on a variety of NLP tasks,} we investigated several deep learning (DL) models for reference detection. Given the small size of our data, we  found that using transfer learning techniques leads to an increase in F1 score by 10\% over the single task learning baselines. However our experiments also show that our CRF tagger attains superior performance. \OUT{, suggesting that the DL  approaches struggle to generalize for the current choice of task and feature selections in our experimental settings.} This is noteworthy as it shows that conventional methods may be better suited for certain tasks and domains for which scarce data is available. (4) Lastly, our reference resolution  model crucially relies on an information-state architecture  \citep{larsson2000information}. It  constantly updates the dialogue state after each conversational turn, to keep track of  the   visualizations on the screen at that instant and information about each of them 
\OUT{on the visualizations currently appearing on the screen and give context for determining the target visualizations.}    Additionally, while our model encodes preference towards more recently added visualizations to the screen, dialogue state allows access to all those currently visible. This  approach boosts accuracy by 6\% over the baseline, in which only the most recent visualization is an eligible candidate referent).
\OUT{(which we found in analyzing our corpus to be the case 77.3\% of the time).} 

We strongly believe in ecologically valid data, which in our case is  multimodal as well. This data is by nature small, in fact tiny as compared to most current datasets. We believe work like ours complements work conducted on large datasets such as Multiwoz \citep{budzianowski2018multiwoz}, which are somewhat artificially generated. 

\section{Related Work}
{\bf Multimodal Dialogue Corpora.} Efforts to build corpora with referential cues in situated multimodal dialogue are not new (e.g., referential annotations based on speech and eye gaze for puzzle pieces \citep{iida2011multi}; speech and haptic gestures for human-robot interaction \citep{chen2012co,chen2015roles}, and so on). Specifically for visualization, Quda \citep{fu2020quda} includes task annotations for independent queries, but  no dialogue corpus is  available that we are aware of.\\
\noindent {\bf Mutimodal Reference Resolution.} When processes  take place in a real or simulated world, then the user refers to objects not only through language, but also through nonverbal cues such  as gestures \citep{navarretta2011anaphora,qu2008beyond,landragin2006visual,eisenstein2006gesture}, eye gaze \citep{prasov2008s,iida2011multi,kim2017multi}, and haptic information \citep{foster2008roles,chen2015roles}. The objects being referred to could be text entities introduced earlier in the discourse or  those in external environments, such as icons on a screen \citep{kehler2000cognitive}, ingredients or tools for a recipe \citep{whitney2016interpreting,chen2012co}.\\
%, and in the kitchen finding useful objects (e.g., \textit{"spoon"}) in specific locations (e.g., \textit{"drawer"}) \citep{chen2012co}. } 
\noindent {\bf Reference Resolution applied to Visualization.} Other visualization systems either limit interaction to system-initiative dialogue flow \citep{cox2001multi}, only handle referents to objects within the current visualization \citep{sun2010articulate, gao2015datatone, narechania2020nl4dv}, or only track referents for follow-up queries on a current visualization \citep{reithinger2005miamm,setlur2016eviza,hoque2017applying,srinivasan2017orko}. Similar to our work, Flowsense \citep{yu2019flowsense} and Articulate \citep{sun2010articulate} are capable of displaying multiple visualizations to the user as well. However in contrast to these applications,
%as first studied more broadly by \citep{webber-baldwin1992}, 
we focus on reference resolution within a dynamic environment \citep{webber-baldwin1992} in which each new visualization introduced into discourse is constructed at that time and can subsequently be moved or removed from the screen at a later time. %While this is not new,
\OUT{ there are scarce publications so far with only a few including document \citep{wilson2016table} and dialogue \citep{li2016reference}, and none in the visualization domain. } Furthermore, we are the first (as far as we know) to propose an ML approach to identifying visualization referring expressions compared to  rule-based techniques \citep{hoque2017applying}. \\
%used currently (e.g., Evizeon \citep{hoque2017applying} assumes the RE along with any data attributes are in the same phrasal chunk however that is not guaranteed with free-forming NL).
\noindent {\bf Transfer Learning techniques} are known to boost deep learning performance in low resource settings such as ours. \OUT{(\textit{CITY-CRIME-VIS} is comprised of only 3.2K utterances compared to popular dialogue corpora such as SwDA \citep{jurafsky1997switchboard} with 193K utterances and MRDA \citep{shriberg2004icsi} with a 78k utterances).} Transfer learning gives a model new insight from auxiliary tasks rather than training from scratch on the target task, consequently reducing training time and improving generalization on smaller data \citep{pan2009survey}. It is a vital tool for NLP due to data scarcity for domain focused language data, gaining considerable popularity especially for contextual embeddings \cite{rahman2020integrating,howard2018universal,radford2018improving},
\OUT{e.g., Bert \citep{devlin2018bert, rahman2020integrating}, ULM-Fit \citep{howard2018universal}, GPT \citep{radford2018improving}, while also successful in other applications such as} speech recognition \cite{song2019topic}, sequence tagging \citep{perl2020low,sogaard2016deep}, and user adaptation \citep{genevay2016transfer}. 
%spoken dialogue systems 

\section{Dataset and annotations} \label{sec:corpus} Our \textit{City-Crime-Vis} corpus  comprises  multimodal interaction for 16 subjects that explored public crime data in our city to better deploy police officers.  As noted, they spoke with a human VE who remotely created visualizations on a large screen, was not visible and did not speak  back. The corpus contains 3.2K utterances.
Since the user was encouraged to reason  out loud about the \OUT{current state of} patterns discovered via visualization, conversational turns often start  with \textit{think aloud}, followed by what we called an  \textit{actionable request} (AR) for the VE. 
\OUT{In turn the VE would respond accordingly with updates to displayed (visualization) content on the screen. } 

\begin{figure*}[htb]
\centering
\includegraphics[scale=0.5,keepaspectratio]{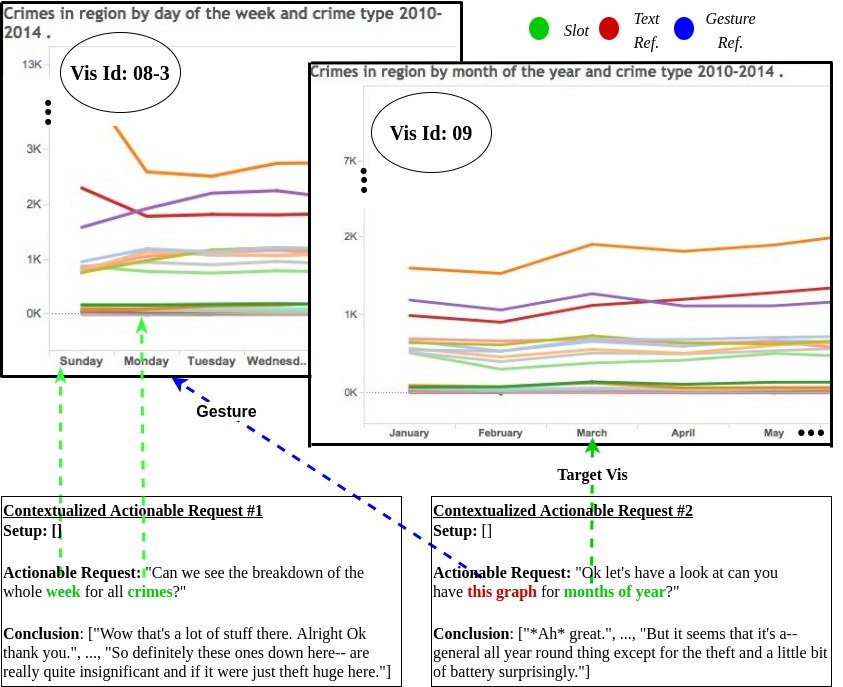}
\caption{Excerpt comprising two  Contextualized Actionable Requests (CARs); references  shown in red (text) and blue  (gesture), and slot fillers in green. In  CAR \#1  future visualization \textit{"08-3"} is specified via temporal axis DAY associated with slot filler \textit{"week"} and similarly CRIME for \textit{"crimes"}. In CAR \#2,  follow-up request to \textit{"08-3"} (referred to via text reference \textit{"this graph"} and co-occurring gesture)  creates  \textit{"09"} substituting  temporal  axis DAY with MONTH, associated with slot filler \textit{"month of year"} (identifiers are for clarity but not part of  display).}
\label{fig:dialogue_annotation_sample}
\end{figure*}

Using ANVIL \citep{kipp2001anvil,kipp2014anvil}, we annotated 449 utterances in context as ARs, hence obtaining 449 CARs (\textit{contextual actionable requests}): a CAR consists of \textit{setup}, i.e. think aloud prior to the AR (up to and including utterances that mention data attributes, if any); the AR; and the \textit{conclusion}, the think aloud subsequent to the AR (also based on data-attribute mentions). %
%In total, this amounts to about 1.5k utterances belonging to a CAR. 
Figure~\ref{fig:dialogue_annotation_sample} illustrates an excerpt  from the corpus, comprising two CARs. 
%(full details in Sec.~\ref{sec:referring_expressions}
%(see Section \ref{sec:referring_expressions} for more detail).}

\OUT{We have made inroads with the data, showing language understanding improves in performance when informed by the think aloud (we marked the AR along with think aloud surrounding it as contextual (calling it \textit{contextual actionable request}) (CAR)). In particular, we showed the benefits of nearby think aloud in informing the user request for determining user intent and paraphrase-based data augmentation. Figure \ref{fig:dialogue_annotation_sample} illustrates sample dialogue from the corpus (see Section \ref{sec:referring_expressions} for more detail).}

\OUT{Overall, the corpus contains 3.2K utterances of which approximately half we have demonstrated to be informative towards interpretation of ARs (i.e., approximately 1.5K utterances belong to a CAR). In a CAR,  }
Each AR is annotated for user intent with one of 8 Dialogue Acts (DA)  labels, e.g., WINMGMT for window management operations, such as closing, moving, maximizing, or minimizing a visualization;  CREATEVIS for creating a new visualization from scratch;  MODIFYVIS for creating a new visualization based on an existing one. %, and so on. 
Referential expression annotation is described next. 
Full details on the annotation, including intercoder agreement, can be found in (Reference withheld). The transcribed corpus is publicly available;  and so is an augmented dataset built to alleviate data scarcity,  comprising a 10-fold increase to 160 subjects covering approximately 15K utterances obtained via delexicalization and paraphrasing.

\subsection{Referring Expressions}
%Annotation and Analysis}
\label{sec:referring_expressions}
\OUT{In this work, we focus on several kinds of visualization references. This includes any}
We annotated both \textit{text} (NPs) and \textit{gestural}  references to  visualizations. Hand gestures were coded with various labels (e.g., the kind of gesture, the objects pointed to on the screen, and so on); approximately a third were identified as referential when they co-occur with text references.
%(if it is made simultaneously with a text reference) called  \textit{gesture reference}. 
%In terms of determining the target visualizations referred to by text  references, 
%at the time the user describes the visualization they wish to see, 
Within text references, we also identify certain \OUT{words} phrases  as slot fillers corresponding to data attributes (i.e., \textit{slots}) in our knowledge ontology (KO)\footnote{Semi-automatically constructed via external sources comprised of 3.5K total terms categorized into 11 parent slots such as CRIME TYPE, NEIGHBORHOOD, MONTH, YEAR (Reference withheld)}. 
\OUT{Later on, if the user refers back to the visualization verbally, we can build a similar semantic structure for the RE by extracting entities from the utterance and then compare that to the semantic representation of visualizations on the screen to select the most likely target.}

%(see Section \ref{sec:referring_expressions} for more details). 

We labeled a total of 294 references in the corpus (176 text references (60\%) and 118 gesture references (approximately one third co-occur with text references) (40\%)) as well as 680 slot fillers. For reference resolution, we attained an excellent intercoder agreement of $\kappa=0.85$ with 2 judges on  the full interaction from one subject: the transcript and potential references were provided,  and each of the judges filled out the referent target visualization for each reference.

Currently, for simplicity, our model focuses on single references occurring in {\it setup} and {\it AR}, not in {\it conclusion}.
\OUT{evaluation of the reference resolution model to labels corresponding to a \textit{single reference}.} Consider %the representative sample from the corpus, 
\textit{"Can you bring up the the graph behind the \OUT{02-2} River North one?"}. The user refers to two visualizations here: 
%identifies visualization \textit{"02-2"} via referring expression 
the \textit{"the \OUT{02-2} River North one"} and the graph placed behind it on the screen.
\OUT{with referring expression \textit{"the graph"}} We only evaluate \textit{"the graph"} in this case since we only process a single reference per request. We also only consider visualization references to a \textit{single target}. The request \textit{"[...] I would like to see battery by-- well I would like to see battery by day of week, battery by month, and battery by year."} results in 3 new  corresponding visualizations. However, our model only adds one of these visualizations to the dialogue history (DH) as part of the evaluation and hence we ignore any references to either of the other two visualizations, which are not in the DH.

%breaks down the label counts
%in the corpus 
Table~\ref{tab:label_cnts}  presents reference counts for \OUT{the evaluation for} setup and ARs: \OUT{it shows that} 5\% of text references are in  setup and  62\% in ARs; \OUT{it also shows} 
8.5\% of gesture references are  in  setup and 74\% in ARs; hence, 33\% of text references (52) and 18\% of gesture references (21)  appear in  conclusion. Single references account for about 94.7\% of references in  setup  and for about 80\% of those in ARs. Finally, when filtering on single targets, we are left with the  80 text and 54 gesture references on which we will focus.

\begin{table}[t!]
    \begin{center}
    \small
    \begin{tabular}{|l|c|c|}
    \multicolumn{3} {c} {\textit{Text References}} \\ \hline
    \multicolumn{3} {|c|} {\textbf{Total:} 176} \\ \hline
    \textbf{Category} & \textbf{Setup} & \textbf{Request} \\ \hline
    Overall & 19 & 109 \\
    Single References & 18 & 86 \\
    Single Targets & 14 & 66 \\
    \hline
    \multicolumn{3} {c} {\textit{Gesture References}} \\ \hline
    \multicolumn{3} {|c|} {\textbf{Total:} 118} \\ \hline
    \textbf{Category} & \textbf{Setup} & \textbf{Request} \\ \hline
    Overall & 10 & 87 \\
    Single References & 9 & 70 \\
    Single Targets & 9 & 45 \\
    \hline
    \end{tabular}
    \end{center}
    \caption{Text and gesture reference corpus counts.}
    \label{tab:label_cnts}
\end{table}

\OUT{\caption{Dialogue is segmented into Contextualized Actionable Requests (CARs). We highlight the multimodal references including text (red) and gesture (blue) as well as slot fillers and the corresponding entity in our KO (green). First, CAR \#1 shows the user in a visualization construction specifies future visualization \textit{"08-3"} via temporal slot DAY associated with slot filler \textit{"week"} and similarly CRIME for \textit{"crimes"}. Subsequently segment CAR \#2 indicates the user following up on visualization \textit{"08-3"} (via text reference \textit{"this graph"} and co-occurring gesture) with another request to create future visualization \textit{"09"} with slot MONTH associated with slot filler \textit{"month of year"}, consequently changing the temporal horizontal axis from day in \textit{"08-3"} to month in \textit{"09"}.}}

\section{Approach: Detection, Resolution, and New Entity Establishment}

 \label{sec:approach}
\begin{figure*}[htb]
\centering
\includegraphics[scale=0.4,keepaspectratio]{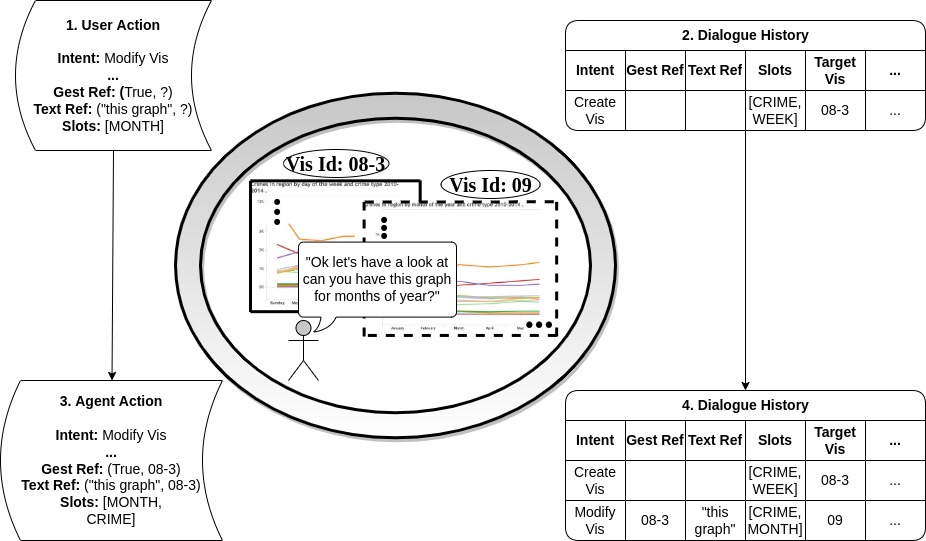}
\caption{The user (inside the circle) currently has visualization \textit{"08-3"} on the screen and is asking to construct "future" visualization \textit{"09"} (in dashed lines).  Language understanding creates user action (1); then the dialogue manager takes dialogue history (2) and creates agent action (3); finally the state tracker updates dialogue history (4).}
\label{fig:reference_resolution_pipeline}
\end{figure*}

The pipeline relies on an information state architecture with dialogue state tracking.
\OUT{where dialogue state tracking plays a central role in giving necessary context to the dialogue manager to form effective responses to the user in conversation.} Reference resolution is carried out in two particular scenarios: the user asks to (1) perform window management on a current visualization (e.g., close, move, and so on) or (2) create a new visualization based on current data or template. An example for the latter is shown in Fig.~\ref{fig:reference_resolution_pipeline}, in which the user asks to construct visualization \textit{"09"} by using \textit{"08-3"} as a template. The pipeline begins with language understanding of the full CAR \#2 from Fig.~\ref{fig:dialogue_annotation_sample} (we only show the AR part of the entire CAR segment here for simplicity) to form a \textit{User Action} frame containing fixed frame attributes. While user intent, visualization references, and slots are filled, others are left empty either because of under-specification by the user (e.g., axes labels, plot title, plot type, and so on) or they require back-end processing (e.g., query results to retrieve requested data, referent visualization identifiers to resolve references). Then the state tracker uses the DH to keep track of which visualizations are on the screen. In Fig.~\ref{fig:reference_resolution_pipeline}, the DH contains a single entry for \textit{"08-3"}. Subsequently, the dialogue manager executes a dialogue policy which aside from making back-end decisions such as forming an SQL query for data retrieval, also seeks to populate the unknown frame attribute values; it outputs an \textit{Agent Action} frame (structurally identical to User Action). Finally the state tracker adds the Agent Action as a new entry in the DH while the system also outputs a json object (which we call a \textit{visualization specification}) that instructs a separate visualization interface software to accordingly update the screen (i.e., add visualization \textit{"09"} in this case). 

As concerns referring expressions and their resolution, the pipeline undertakes the following steps, to be described in detail next: referring expression detection; semantic structure / visualization vector construction;  reference resolution; new entity establishment. 

\OUT{ (BDE this list basically repeats what's in the 4 coming subsections.)
\begin{enumerate}
    \item The referring expression \textit{"this graph"} is detected.
    \item The user intent, determined to be Modify Vis in this case, informs the pipeline that reference resolution is necessary to identify the visualization on the screen to be used as a template for constructing the new visualization.
    \textit{"months of year"} is embedded using a domain targeted word embedding model (WE)\footnote{100-dimensional continuous bag-of-word model trained on 5GB of online articles and wikipedia pages related to crime.} to find the MONTH slot which is the closest slot in the KO. In this case, the slot along with inferred plot type (i.e., line chart for temporal entities) form a semantic representation for the referring expression.
    \item Visualization \textit{"08-3"} is identified as the target visualization via the similarity in semantics between the referring expression and the entries in the DH (when visualization \textit{"08-3"} was constructed in earlier dialogue its semantics were formed in a similar fashion to the referring expression).
    \item Finally visualization \textit{"09"} is constructed using the \textit{"08-3"} template. Note that MONTH serves a double purpose since it was used to resolve the referring expression while also replacing WEEK as the temporal axis in \textit{"09"}.
\end{enumerate}}

 \subsection{Detection}
 \label{sec:approach_detection}
We trained a sequence tagging model to detect text references (see Section \ref{transfer_learning} for details). The model predicts tags using the standard IOB2 format (i.e., \textit{"B-REF"/"I-REF"/"O-REF" } for beginning of / inside / outside text reference respectively.
%\textit{"I-REF"} for inside 
%text reference,  and \textit{"O-REF"} for outside the text reference boundary). 
In the current example, our model tagging output is: [(Ok, O), (let's, O), (have, O), (a, O), (look, O), (at, O), (can, O), (you, O), (have, O), {\bf (this, B), (graph, I)}, (for, O), (months, O), (of, O), (the, O), (year, O)]. The User Action frame (\#1 in Fig.~\ref{fig:reference_resolution_pipeline}) attribute {\bf Text Ref.}  is updated accordingly. %with \textit{"this graph"}.

Gestures are determined to be referential if they are pointing to a visualization \OUT{appearing} on the screen and also co-occur with a text reference.
%in the utterance. 
In our example, a text reference is present (\textit{"this graph"}) hence the User Action frame boolean attribute {\bf Gest Ref.} is set to  \textit{True}.

\subsection{Semantic Structure Construction}
Each time a %new visualization is introduced into the discourse (i.e., when 
user asks to construct a new visualization, %entities in the KO 
our model looks for slots in the request to form the semantic structure for the new visualization (this also applies for the referring expression). In particular, we find phrases that are in close proximity in the embedding vector space to terms in the KO, by using  a domain targeted word embedding model (WE)\footnote{100-dimensional continuous bag-of-word model trained on 5GB of online articles and wikipedia pages related to crime.}.
Subsequently the candidate words are pruned based on linguistic patterns using the SpaCy\footnote{http://spacy.io} dependency parse of the entire utterance to form the finalized list of slot fillers. For example in the  AR  in Fig.~\ref{fig:reference_resolution_pipeline}, 
%the user utterance includes 
the prepositional phrase \textit{”for months of year”} contains the \OUT{prepositional} complements \textit{”month”} and
\textit{”year”}, both of which are known as temporal slots in KO. Here, the terms are merged to form \textit{”months of year”}, and mapped to the parent slot MONTH, since \textit{"month"} appears first in the phrase.

The slot fillers are then transformed to low-dimensional space. In particular, the 11 slots in the KO are projected onto an embedding space along 11 dimensions, each computed using the WE model, representing features for visualization and referring expressions. In case a slot filler corresponds to a slot value in the KO, it is simply averaged into the corresponding parent slot position in the feature vector.

\subsection{Reference Resolution} \label{sec:approach_resolution} 
In Fig.~\ref{fig:reference_resolution_pipeline}, the DH contains only an entry for \textit{"08-3"} introduced previously in the interaction. Subsequently, when resolving visualization references for future visualization \textit{"09"} the model only considers \textit{"08-3"} as a candidate referent (it is selected because their cosine similarity score exceeds a cut-off score).

Our model encodes preference to the most recent entries. If $n$ represents the total entries in the DH, then the visualization vectors of the most recent $\frac{n}{2}$ entries in the DH are associated with a multiplicative factor of 1.0 signifying that they are equally preferred. The latter $\frac{n}{2}$ entries in the DH however are associated with a linear decrease by a factor of $\frac{1}{n}$. For example, if $n=6$, then any of the most recent 3 entries are equally likely candidates; the visualization vectors for the remaining entries are multiplied by factors of $\frac{2}{3}$, $\frac{1}{3}$, and $\frac{0}{3}$ respectively. Finally, cosine similarity is used to score each visualization in the DH relative to the referring expression and the visualization with the highest score is selected.

\subsection{New Entity Establishment}

Finally visualization \textit{"09"} is constructed using the \textit{"08-3"} template. Note that MONTH serves a double purpose since it was used to resolve the referring expression (via WE embedding and cosine similarity among semantic structures), while also replacing WEEK as the temporal axis in \textit{"09"}.

Our model also infers missing information the user does not mention in the request. For example plot type is set to line chart in the presence of temporal entities to better display trends across time. Otherwise, by default bar chart would be selected since 56\% of all visualizations in the corpus are bar charts. In the case of \textit{follow-up requests}, in which the user makes reference to a  visualization from a previous query, information can be added from the original request. In the running example, CRIME is added to the entities list in the Agent Action because \textit{"08-3"} of the previous request includes it. 

\section{Experiments and Results}

\subsection{Detection}
\label{transfer_learning}
As described in Sec.~\ref{sec:approach_detection}, we treat the problem of detecting text references (DTR) as a sequence labeling task. With only 294 references appearing across 449 CARs in the corpus, the task has access to a limited number of labels for model training. We investigated transfer learning to address data insufficiency, i.e. our goal is to transfer knowledge from a source sequence tagging task with access to a large number of labels (i.e., NER in our case) to DTR. The \textit{City-Crime-Vis} corpus data is supplied to the DTR task associated with 3 possible tags (see Sec.~\ref{sec:approach_detection}) while the augmented dataset of approximately 15K utterances (see Sec.~\ref{sec:corpus}) is provided to the NER task,  comprising 23 labels (the \textit{"B"} and \textit{"I"} tag for each of the 11 parent slots in the KO plus \textit{"O"} tag). 

For knowledge transfer in sequence tagging problems, a generalizable architecture is preferred to optimize relatedness of tasks to maximize weight sharing \citep{collobert2011natural}. Advanced approaches known to combine the benefits of CRF with deep learning \citep{collobert2011natural,hammerton2003named,jagannatha2016structured,chiu2016named}  in particular can be leveraged where only the CRF layer needs to be adjusted to account for the difference in labels while sharing the other layers \citep{DBLP:journals/corr/YangSC17}. For our experiments, we chose CRF for sequence tagging  \citep{fields2001probabilistic,sha2003shallow} which is implemented using Sklearn-CRF Suite\footnote{https://sklearn-crfsuite.readthedocs.io/en/latest/} package; and also chose BiLSTM-CRF and BERT-CRF are implemented in Keras\footnote{https://keras.io/}.
\subsubsection{Features}
The input to the models consist of utterances (20 words maximum), coarse-grained POS tags, and sequential text features. They are trained using 5-cross validation. For training CRF we used L-BFGS optimization with 0.10 coefficients for L1 and L2 regularization. BiLSTM-CRF is trained on a batch size of 1024 and 10 epochs and BERT-CRF is trained on batch size of 64 and 5 epochs. Early stopping was applied to F1 score to address class imbalance. Finally,  \textit{"O"} tags which commonly dominate other labels in NER data are addressed by skipping utterances only containing \textit{"O"} labels. The following describes the deep learning models:
%in detail.
\begin{enumerate}[label=(\alph*)]
    \item \textbf{Single Task learners (Baselines)} BiLSTM-CRF consists of an embedding layer of 100 hidden units followed by a BiLSTM layer for the current utterance and similarly for coarse-POS tags, followed by a merge layer and dense layer and finally a CRF layer. The BERT-CRF architecure has a  similar architecture to BiLSTM-CRF except it uses a BERT layer (corresponding to the pre-trained cased model) to output 768 dimensional contextual embeddings for the utterance input and  an embedding layer and BiLSTM layer to process POS tag input. Both models are trained on the target DTR data.
    \item \textbf{Sequential Transfer Task Learners (STL)} The models are trained on the NER data first using a CRF layer with 23 output dimensions to build pre-trained models. Then, the CRF layer is replaced with a new untrained CRF layer with 3 dimensions output for fine tuning with the DTR data.
    \item \textbf{Multi Task Learners (MTL)} In both BiLSTM-CRF and BERT-CRF, the model architecture is altered to share weights from all the layers prior to the CRF layer between the NER and DTR tasks. The CRF layer cannot be shared due to difference in number of possible labels for the two tasks. 
\end{enumerate}

\noindent %\textbf{Weight sharing.} No need for this title now. 
Results in table \ref{tab:transfer_learning_results} show that transfer learning boosts the performance of the deep learning models. In particular, the multi-task learners (i.e., BiLSTM-CRF-MultiL and BERT-CRF-MultiL) statistically significantly \footnote{One-way Anova with post-hoc Tukey HSD.} ($p<0.001$) outperform the baselines (i.e., BiLSTM-CRF, BERT-CRF) by over 10\% F1. This is consistent with the literature which indicates weight sharing allows increased sharing of knowledge between similar neural network architectures.

\begin{table}[t!]
    \begin{center}
    \small
    \begin{tabular}{|l|c|}
    \hline \textbf{Model} & \textbf{F1} \\ \hline
    BERT-CRF & 0.318 \\
    BiLSTM-CRF & 0.327 \\
    BERT-CRF-STL & 0.345 \\
    BiLSTM-CRF-STL & 0.404 \\
    BiLSTM-CRF-MTL & 0.435 \\
    BERT-CRF-MTL & 0.435 \\
    CRF & \underline{0.616} \\
    \hline
    \end{tabular}
    \end{center}
    \caption{Text ref. detection using transfer learning. CRF  statistically significantly outperforms other models.}
    \label{tab:transfer_learning_results}
\end{table}

\noindent %\textbf{CRF Tagger.} 
Table~\ref{tab:transfer_learning_results} also indicates CRF performs statistically significantly better than all other models. This suggests that auxiliary tasks in addition to NER may be necessary to further see improvement as well as adding features other than POS tags.
This reaffirms our choice for the detection framework (i.e., CRF), which  we use  in the  evaluation of  the whole reference resolution pipeline in the following. Note that only detection can benefit from deep learning / transfer learning, since it can be set up as a classification problem.
%\section{Results}
\subsection{Reference Resolution Pipeline}
\noindent \textbf{Detection.} 
 Table \ref{tab:ref_detection_results} shows how accurately our model detects the annotated labels for the reference resolution task; recall there are a total of 54 gesture and 80 text references cross setup and request, achieving an overall accuracy score of 52.2\%. Note that the model performed worst on gesture references. One possible reason for this is that the heuristic rule for deciding whether a gesture is referential may need to be relaxed (we currently ignore any gesture that did not co-occur with a text reference). Another factor could be that our CRF tagger model for detecting text references may need further exploration potentially taking gesture features into account since they often co-occur.
 
 \begin{table}[t!]
    \begin{center}
    \small
    \begin{tabular}{|l|c|c|}
    \hline \textbf{Label} & \textbf{Setup} & \textbf{Request} \\ \hline
    Text Ref. & 60.0 & 55.0 \\
    Gesture Ref. & 53.0 & 39.0 \\
    \hline
    \end{tabular}
    \end{center}
    \caption{Accuracy on detecting referring expressions (B-REF, I-REF, O-REF).}
    \label{tab:ref_detection_results}
\end{table}

\begin{table}[t!]
    \centering
    	\begin{tabular}{|l|c|c|c|c|c|}
    		\hline
    		& \multicolumn{5} {c|} {\textbf{Slots Detected}}\\
    		Type & $=0$ & $\leq25$ & $\leq50$ & $\leq75$ & $\leq100$ \\ \hline
    		 AR & 16 & 5 & 18 & 67 & 132 \\
    		 FR & 16 & 0 & 4 & 8 & 13 \\ \hline
    	\end{tabular}
        \caption{Total ARs with percent of slots matched. Also shown for follow up (FR), a small subset of AR.}
        \label{tab:entity_detection_quartile_results}
\end{table}
 
\noindent \textbf{Semantic Structure Construction.} Overall, our model achieved an accuracy score of 66.2\% for detecting slots as part of the new entity establishment process for visualization features. Further analysis from Table~\ref{tab:entity_detection_quartile_results} shows that in over half of all utterances (57.6\%), all the slots were detected while in 85.3\% of the utterances, at least 75\% of them were successfully detected. We will discuss follow-up requests shortly.
 
\noindent \textbf{Resolution.}
Accuracy  on resolving text and gesture references for varying \textit{WINDOW} sizes is shown in Table~\ref{tab:reference_resolution_results}.  Focusing first on unlimited window size (i.e., $\infty$ for all referent visualization candidates are eligible), we see 100\% accuracy for gesture references for both setup and request as expected: ground truth gestures are used for input for the experiment. This also validates that our DM is correctly performing state tracking. Next, text references found in ARs are correctly resolved for over 68\% while for setup it is 85\%. Overall accuracy score for resolving references is  83.6\%.
\begin{table}[t!]
    \begin{center}
    \begin{adjustbox}{max width=0.85\textwidth}
    	\begin{tabular}{|l|c|c|c|c|c|c|}
    		\hline
    		& \multicolumn{3} {c|} {\textbf{Setup Win. =}}
    		& \multicolumn{3} {c|} {\textbf{AR Win. = }} \\
    		%    \cline{1-1}
    		Ref & 0 & 1 & $\infty$ & 0 & 1 & $\infty$ \\ \hline
    		  Gest. & 0.0 & 93.0 & 100.0 &  0.0 & 68.4 & \underline{100.0}\\ \hline
    		  Text &0.0 & 85.3 & 85.3 & 0.0 & \underline{74.4} & 68.3  \\ \hline
    		 All & 0.0 & 82.6 & 87.0 &0.0 & 73.0 & \underline{82.9} \\\hline
    		 
    	\end{tabular}
    \end{adjustbox}
    \end{center}
\caption{Resolution accuracy for varying window sizes.}
\label{tab:reference_resolution_results}
\end{table}

Turning now to other window sizes,  an expected trend can be observed in Table~\ref{tab:reference_resolution_results}: as one increases the number of eligible candidate referents for resolution, there is an overall increase in correctly resolved visualization references. In particular  dialogue context boosts accuracy by approximately 10\% for ARs and 4\% for setup. However, text references struggle with increasing window size, suggesting our linear decay function may need further tuning to better model the user preference behavior. Further analysis actually shows similar behavior between the model and user: in over 75\% of the time the most recent visualization is chosen by both, however for further away entries the linear decay function penalizes them too harshly leading to the model rarely choosing them.

\OUT{
\noindent \textbf{New Entity Establishment.} Overall, our model achieved an accuracy score of 66.2\% for detecting slots as part of the new entity establishment process for visualization features. Further analysis from Table~\ref{tab:entity_detection_quartile_results} shows that in over half of all utterances (57.6\%), all the slots were detected while in 85.3\% of the utterances, at least 75\% of them were successfully detected. We will discuss follow-up requests shortly.}

\noindent \textbf{Under-specification.} Table \ref{tab:entity_detection_quartile_results} shows in over half the cases (51.2\%), at least 75\% of slots in follow up requests are detected by our model. Moreover  Table~\ref{tab:underspecified_entity_results} shows that the model achieves close to a 6\% boost in accuracy, however  the model struggles to identify slots beyond one request in follow up sequences of requests. Next, Table~\ref{tab:underspecified_plot_type_prediction_results} shows model performance on plot type prediction. The model predicts between three kinds of plots including heat map, line chart, and bar graphs. Bar graphs achieved superior performance. Over half the visualizations (56\%) shown to the user are bar graphs. Heat maps performed worst; it is not always obvious when to use heat maps. For instance, asking for crimes by neighborhood could be shown as a bar graph to quickly determine the safest and most dangerous neighborhoods whereas as a heat map the location of the neighborhoods is also shown.

\begin{table}[t!]
    \begin{center}
    \small
        \begin{tabular}{|l|c|}
            \hline \textbf{Plot type} & \textbf{F1} \\ \hline
            Heat map & 0.467 \\
            Line chart & 0.511 \\
            Bar graph & 0.739 \\ \hline
        \end{tabular}
`    \end{center}
    \caption{Average F1 score for predicted plot type.}
    \label{tab:underspecified_plot_type_prediction_results}
\end{table}

\begin{table}[t!]
    \begin{center}
    \small
        \begin{tabular}{|l|c|}
            \hline \textbf{WINDOW} & \textbf{Acc} \\ \hline
            0 & 60.0 \\
            1 & 65.8 \\
            $\infty$ & 66.2 \\ \hline
        \end{tabular}
    \end{center}
    \caption{Slot detection accuracy for varying window sizes.}
    \label{tab:underspecified_entity_results}
\end{table}

\section{Conclusions and Future Work}
We have presented a reference resolution model for multi-modal visualization dialogue. In particulate, the model resolves visualization references in the context of the current interaction, crucially tracking visualizations constantly being added and removed from the screen. Also the model is central to the  creation of new visualizations: visualization features via entity introduction help the model know how to refer to a visualization later on. 

We plan to address potential extensions in future work. First, the model assumes proximity to measure the word dependency relationship between a referring expression and nearby slots (e.g., \textit{"months of year"} is in the same utterance as \textit{"this graph"}). Incorporating additional linguistic information via the dependency parse tree can capture this relationship more reliably.  Other worthwhile avenues to explore include ways to better model user behavior for referring to more distant visualizations and  adapting our resolution algorithm beyond the cosine similarity measure to more sophisticated machine learning based approaches to take advantage of the rich visualization feature space in our case. Also, as part of our broader research objectives, we intend to conduct studies with real users  interacting with our visualization system, by integrating our resolution pipeline with the first version of the assistant: there, we used an Android app that employs the Google Speech API to perform speech-to-text on the click of a button and a gesture recognition system that recognizes pointing gestures using a Microsoft Kinect camera calibrated for our large screen display.
\OUT{we plan to interface with multimodal inputs including an Android app that uses the Google Speech API to perform speech-to-text on the click of a button and a gesture recognition system that recognizes pointing gestures using a Microsoft Kinect camera calibrated for our large screen display.} 


\bibliography{custom}
\bibliographystyle{acl_natbib}

% \appendix
% \section{Frequency of Preferred Referent}
% \begin{figure*}[htb]
% \centering
% \includegraphics[width=2\columnwidth,height=2\textheight,keepaspectratio]{Figures/freq_preferred_referent.png}
% \caption{Frequency of how often the user refers to a specific entry in the DH. Model refers to the linear decay function and User corresponds to actual user behavior.}
% \label{fig:freq_preferred_referent}
% \end{figure*}

\end{document}